\documentclass{article}
\usepackage{spconf,amsmath,graphicx}
\usepackage{multirow}

\input{./Definitions}
\renewcommand{\paragraph}[1]{\noindent\textbf{#1}\quad}

\title{Correction Focused Language Model Training for Speech Recognition}
\name{Yingyi Ma, Zhe Liu, Ozlem Kalinli}
\address{Meta AI, Menlo Park, CA, USA}
\begin{document}
\ninept
\maketitle
\begin{abstract}
Language models (LMs) have been commonly adopted to boost the performance of automatic speech recognition (ASR) particularly in domain adaptation tasks. Conventional way of LM training treats all the words in corpora equally, resulting in suboptimal improvements in ASR performance. In this work, we introduce a novel correction focused LM training approach which aims to prioritize ASR fallible words. The word-level ASR fallibility score, representing the likelihood of ASR mis-recognition, is defined and shaped as a prior word distribution to guide the LM training. To enable correction focused training with text-only corpora, large language models (LLMs) are employed as fallibility score predictors and text generators through multi-task fine-tuning. Experimental results for domain adaptation tasks demonstrate the effectiveness of our proposed method. Compared with conventional LMs, correction focused training achieves up to relatively 5.5\% word error rate (WER) reduction in sufficient text scenarios. In  insufficient text scenarios, LM training with LLM-generated text achieves up to relatively 13\% WER reduction, while correction focused training further obtains up to relatively 6\% WER reduction.
\end{abstract}
\section{Introduction}
\label{sec: intro}
In the past decade, neural network language models (NNLMs) have gained widespread success within state-of-the-art automatic speech recognition (ASR) systems, primarily through the first-pass fusion \cite{KanWuNgu2018, KimShaMah2021, ChaJaiNav2016} and second-pass rescoring \cite{IriZeyAlb2019, LiuWanChe2014, XuCheGao2018, LiPovKhu}. Particularly, NNLMs play an important role in domain adaptation tasks due to their strong adaptability on text-only corpora for target domains.

Traditional methods for training LMs treat all the words in a sequence equally, aiming at minimize the perplexity (PPL). While reducing PPL can lead to higher likelihood on ground truth words in the references, a substantial portion of these words may already be correctly recognized by the ASR systems without external LMs. Consequently, improving PPL for such words does not necessarily contribute much to reducing the word error rate (WER). On the other hand, LM capacity might be constrained by runtime requirements \cite{RajFilTiw2019, NusVerOua2023}, rendering it less cost-effective when treating all words equally. To address these challenges in domain adaptation tasks, our attention shifts towards ASR fallible words during LM training, and we aim to develop strategies of incorporating ASR fallibility information into LM training corpora.

References annotated with word errors are valuable resource for accessing ASR fallibility, where paired data of utterances and references are typically required. In many real-world domain adaptation scenarios, however, we might only be able to acquire a small amount of such pairs from the target domains. Additionally, a small to moderate amount of text-only records may be available, which are useful for conventional language modeling but still lack the ASR fallibility annotations that we need.

It thus calls for an advanced approach that can learn from a small number of utterance-reference pairs and is then capable of predicting ASR fallibility information for text-only corpora. These predictions should well reflect the ASR performance on the target domains. Furthermore, in cases where text-only data is insufficient, this method is able to generate target-domain text records along with the predicted fallibility measurements. Recently, large language models (LLMs) are widely recognized as remarkable few-shot learners and text generators, which could be promising for these purposes \cite{MenMicHua2023, GaoPiYon2022, GaoFisChe2020}. 

In this paper, we present a new correction focused NNLM training strategy. Specifically, we introduce ASR fallibility score, which quantifies the likelihood of a word being mis-recognized by a given ASR system. The score is at the word level and context dependent. The we propose to leverage LLMs as both fallibility score predictors and text generators through multi-task fine-tuning. Armed with a rich text-only corpus and the associated fallibility scores, we employ the se scores to form a prior word distribution. Our proposed approach prioritizes LM training towards ASR fallible words. We conduct experiments on domain adaptation tasks, considering both sufficient text and insufficient text scenarios. Our results show that LMs resulted from our proposed correction focus training provides profound improvements compared with conventionally trained LMs.

\section{Related Works}
\label{sec: related_works}
Discriminative training techniques have been extensively explored in the field of speech recognition. Discriminative loss such as minimum WER (MWER) is widely adopted in training second-pass rescoring models, with the aim to favor hypotheses with the lowest WER \cite{XuGuKol2022,PraSaiWu2018,MenWuKan2021,WeiCheSai2022}. However, in latency-sensitive systems, there is a preference for utilizing LMs in first-pass fusion. Recently, authors in \cite{FutInaUeo2022} introduce the concept of leveraging masked LMs (MLMs) for error correction within CTC-based ASR systems. Yet, extending this approach to first-pass fusion with RNN-T based models is challenging due to the non-autoregressive nature of MLMs. Furthermore, the discriminative loss employed in these approaches typically require hypotheses lists derived from ASR decoding results for training. Obtaining such information is not feasible when dealing with text-only corpora in domain adaptation tasks.

In the absence of ground truth references for ASR transcripts, confidence scores have been utilized as an important metric to assess the token-wise reliability of ASR hypotheses \cite{Jia2005, HuaKumLiu2013, KalLiuGon2015}. When ASR transcripts serve as the primary text resource for training LMs, confidence scores can be employed for data selection, allowing for filtering out the transcripts with low reliability \cite{XieChe2013,LiuLiBak2021}. However, confidence predictors can only assess token-wise reliability based on ASR decoding results, thus are hard to help the scenarios where text-only data are more abundant than ASR transcripts. This necessitates a predictor that can estimate the likelihood of ASR errors for tokens with only text input. Furthermore, while confidence scores aid in selecting confident ASR transcripts for LM training, our proposed correction focused training approach places greater emphasis on the tokens where ASR is more likely to make mistakes.

Data sampling and weighting methods have been well studied in the broad deep learning community \cite{FerDow2018, RenZenYan2018, ShuXieYi2019, MaLiuZha2023}. Dynamic reweighing strategies usually requires second-order derivative computation which could be of high cost. Other sampling and weighting approaches for training LMs are mostly towards the direction of minimizing perplexity for target domain corpus. However, the primary goal of training LMs for ASR is to help reduce the WER. Therefore, a word-level focused strategy towards addressing possible ASR mis-recognitions is more appropriate.

Leveraging LLMs in ASR systems has gained significant attention recently. Common approaches of integrating LLMs in ASR as second-pass rescorers \cite{HuSaiLi2023, UdaSuzKur2022, CheAllHua2023} have shown notable improvement on domain adaptation tasks and multilingual ASR systems. Nevertheless, the high cost of inference with LLMs hinders the practical adoption of such approaches. Other branches of research leveraging LLMs to transfer knowledge at ASR model training stage to guide the training of token embedding layer or the whole internal predictor \cite{KubKarBac2022, HanCheShi2023}. However, they neglect the few-shot learning power of LLMs neither focusing on ASR correction. Instead, we propose to leverage LLMs for capturing the ASR fallibility and target domain information with lightweight fine-tuning on limited amount of data. Then such knowledge is transfered to NNLM training through score prediction and text generation. 

\section{Methods}
\label{sec: methods}
In this section, we define fallibility scores on real or generated target-domain text, and illustrate the training approach with more focus on ASR fallible words with associated fallibility scores.

We consider a domain adaptation task where there is a mismatch between the source and target domains. Suppose we have the access to a small amount of utterances paired with their references from the target domain, denoted as $D_{dev}=\{(\ab, \xb)_i\}_{i=1}^{N_d}$, where $\ab$ refers to an audio sequence, and $\xb$ is the associated reference text sequence. We may optionally have a moderate amount of text-only records from the target domain, denoted as $D_{text}=\{(\zb)_i\}_{i=1}^{N_t}$. An NNLM can be trained on $D_{text}$ for helping boost ASR performance on the target domain. 
\subsection{Fallibility Score}
The fallibility score of a word is defined as the likelihood of its being mis-recognized by a given ASR system. Since the ASR may perform differently on the same word when it appears in different sequences, the fallibility score is context dependent.
Given an ASR $\Mcal$, and an aligned audio-text sequence $(\ab, \xb) = (\{a_u\}_{u=1}^{U}, \{x_t\}_{t=1}^{T})$, the ASR likelihood score on word $x_t$ can be derived as $T_{u,t} = P_{\Mcal}(x_t|\xb_{<t}, \ab_{<u})$, the fallibility score of any word $x_t \in \xb$ is then defined as:
\begin{align}
\label{formula:def_score}
 S_{x_t} &= 1-P_{\Mcal}( x_t|\xb_{<t}) \\
 & = 1- \sum_{\ab_{<u}} T_{u,t}\cdot P(\ab_{<u}|\xb_{<t})
\end{align}

Apparently, the computation of fallibility score requires the distribution on $\ab$ and we do not posses that for text-only corpus. We thus need an ASR fallibility score predictor based on the ASR decoding results on the small target domain corpus. 
\subsection{LLM as Fallibility Score Predictor}
We fine-tune an LLM on the paired utterance-reference set $D_{dev}$ using the token classification task. The fine-tuned LLM can then be utilized as a fallibility score predictor. We describe the process into the following steps.

\paragraph{Annotate Paired Corpus.} We first conduct word-level annotation on sequences from $D_{dev}$ with ASR decoding results. An ASR usually makes three types of mistakes: substitution, deletion, and insertion. We annotate each word in $\xb \in D_{dev}$ with the  following rules: (1) If a substitution or deletion error occurs, we annotate the victim word as an {ASR error}; (2) If an insertion error occurs, we annotate the word after the insertion as an {ASR error} (including the \texttt{<eos>} token as well). 
During tokenization, we assign the same annotation label for all the sub-words from the same word. Eventually, we obtain the annotated paired corpus denoted as 
\begin{align}
\label{formula:label_dev}
 \Dhat_{dev} = \{(\xb, \yb)_i\}_{i=1}^{N_d}, \quad y_j=0,1 \quad \forall y_j \in \yb
\end{align}
where $\xb$ is slightly abused to denote the tokenized text sequences as well, $\yb$ represents the annotation sequences carrying the token-level binary {ASR error} labels, where 1 refers to the presence of an error.

\paragraph{Fine-tune LLM.} With the annotated sequences, we fine-tune an LLM so that it can predict if a token is an {ASR error}. We apply a token classification head $\omega$ on top of the main model component of an LLM $\phi$. We can obtain the fined-tuned $LLM_{\phi, \omega}$ via the following optimization objective:
\begin{align}
\label{formula:predictor_loss}
\argmin_{\phi, \omega} \sum_{(\xb, \yb)\in \Dhat_{dev}} -\frac{1}{|\yb|}\sum_{j=1}^{|\yb|} \log P_{\phi, \omega} (y_j|\tilde{\xb})
\end{align}
where $\tilde{\xb}=\xb$ for non-autoregressive and $\tilde{\xb}=\xb_{\le j}$ for autoregressive LLMs. The predicted probability of being an {ASR error} will be served as the fallibility score. We denote the fallibility score facilitated text corpus for $D_{text}$ as below:

\begin{align}
\label{formula:text_set_w_score}
\Dhat_{text} = \{(\zb, \sbb)_i\}_{i=1}^{N_t}, s_j = P_{\phi, \omega}(y_j=1|\tilde{\zb})\quad \forall s_j \in \sbb
\end{align}

\paragraph{Assemble Token-level Fallibility Scores.}
The fallibility scores are predicted at the token level of LLM. However, to enable first-pass fusion with ASR, NNLMs may employ a different tokenizer. Therefore, we need to reassemble the fallibility scores into word level and then split them into NNLM token-level ones. We assign the \textit{max} of the sub-words scores as the word-level fallibility score.

\subsection{LLM as Text and Fallibility Score Generator}
The fallibility score prediction in previous section depends on the availability and sufficiency of $D_{text}$. In the case where more text records are needed, we can fine-tune an LLM to generate in-domain text. Considering the high cost of performing LLM inference, to avoid an additional inference step for fallibility score prediction, we propose the multi-task fine-tuning strategy for LLM to generate text records and predict the fallibility scores simultaneously.
Specifically, we adopt multiple heads in LLM fine-tuning: an LM head $\gamma$ for generation task trained via maximizing the likelihood of each token conditioned on previous tokens, and a classification head $\omega$ for fallibility score prediction. A multi-heads $LLM_{\phi, \gamma, \omega}$ can be obtained through optimizing the following multi-task objective:
\begin{equation}
\begin{aligned}
\label{formula:multitask_loss}
\argmin_{\phi, \gamma, \omega} \sum_{(\xb,\yb) \in \Dhat_{dev}} -\frac{1}{|\xb|} &\sum_{j=1}^{|\xb|} ( \log P_{\phi, \gamma} (x_j|\xb_{<j}) \\
&+\lambda \log P_{\phi, \omega} (y_j|\xb_{\le j}) )
\end{aligned}
\end{equation}
where the hyperparameter $\lambda$ controls the weight for the token classification task. Notice that at every token step, LM head is maximizing the likelihood of the next token while classification head is predicting the score for the current token. A synthetic text corpus from target domain with fallibility scores can then be expressed as: 
\begin{align}
\label{formula:gen_set_w_score}
\Dhat_{gen} &= \{(\zb, \sbb)_i\}_{i=1}^{N_g}\\
\zb &= \{x_{j} = \argmax_e P_{\phi, \gamma}(e|\zb_{< j})\}_{j=1}^{|\zb|} \\
\sbb &= \{ s_j = P_{\phi, \omega}(y_j=1|\zb_{\le j})) \}_{j=1}^{|\zb|}
\end{align}

\subsection{Correction Focused NNLM Training}
A conventional LM is trained with the objective $\max P_{\theta}(\zb|\zb_{<<1})$, where $\zb_{<<1}$ denotes one-step left-shifted $\zb$. To have an NNLM focus more on the ASR fallible words or tokens, we define a prior distribution $Q(\zb)$ where each token $z_j \in \zb$ has its own distribution $Q(z_j)$. To guide the probability distribution of NNLM $P_{\theta}(\cdot)$ to match the prior distribution $Q(\cdot)$, we adopt Kullback–Leibler (KL) divergence. The goal of training an NNLM then becomes minimizing $KL(Q||P_{\theta}) = \EE_Q{\log\frac{Q}{P_{\theta}}}$, which is equivalent to minimizing the following objective:
\vspace{-3pt}
\begin{align}
\label{formula:kl_loss}
 \min_{\thetab} \sum_{(\zb, \sbb) \in \Dhat} -\frac{1}{|\zb|}\sum_{j=1}^{|\zb|} Q(z_j)\log P_{\theta}(z_j|\zb_{< j})
\end{align}
where $Q(z_j) = s_j$, $\Dhat$ denotes either $\Dhat_{text}$ or $\Dhat_{text} \cup \Dhat_{gen}$ depending on the sufficiency of text-only corpora from target domains. To better control the focus intensity over tokens, we introduce a temperature hyperparameter $\alpha \ge 1$, and the correction focused NNLM training objective then becomes:
\vspace{-3pt}
\begin{align}
\label{formula:correction_loss}
 \min_{\thetab} \sum_{(\zb, \sbb) \in \Dhat} -\frac{1}{|\zb|}\sum_{j=1}^{|\zb|} \alpha^{s_j}\log P_{\theta}(z_j|\zb_{< j})
\end{align}
With $\alpha > 1$, tokens with higher fallibility score are assigned higher weights. This encourages NNLM training to focus more on such ASR likely mis-recognized tokens. When $\alpha = 1$, the objective is the same as the conventional training.

In the scenarios where the annotated corpus $\Dhat_{dev}$ from the target domain is not available, we may resort to some annotated corpus from other domains and use it as out of domain for score prediction. The intuition is that ASR may make similar mistakes across different domains such as pronunciation confused words and grammar related errors. In practical applications, sequences are not necessarily annotated with all deletion, insertion, or substitution errors. In the cases of fine-tuning with out of domain $\Dhat_{dev}$, the deletion and insertion errors may be distributed differently from the target domain, we thus optionally tag those errors.

For LLM fine-tuning, the main model $\phi$ is not necessarily to be fully fine-tuned on all parameters. We can adopt parameter-efficient fine-tuning methods such as low-rank adaptation on partial parameters \cite{HuSheWal2022}. For score prediction task, freezing pretrained $\phi$ makes no significant difference on the predicted fallibility scores. For multi-task generator, it is necessary to fine-tune $\phi$ to ensure text generation quality.

\section{Experiments}
\label{sec:exp}
\subsection{Datasets}
We summarize the datasets and their splits in Table~\ref{table:data_summary}. Notice that the statistics in Table~\ref{table:data_summary} only refer to the portions of data used in our experiments. They consist of
\begin{itemize}
\item Publicly available datasets, including Wall Street Journal corpus (\textsl{wsj}), WikiText-103 (\textsl{wiki}) and \textsl{dev} split of Librispeech \textsl{lib-dev}. For \textsl{wsj} corpus, we randomly sample 10K records from \textit{si284} split to serve as the text-audio paired set, then consider \textit{nov93dev} and \textit{nov92} as the test sets. The text-audio set of \textsl{lib-dev} refers to the union of \textit{dev-clean} and \textit{dev-other} splits; 
\item In-house datasets, including a video corpus (\textsl{video}) sourced from public social media videos and a conversation corpus (\textsl{conv}) acquired through crowd-sourcing via mobile devices from a data supplier. Prior to transcription, all of these datasets undergo a de-identification process, ensuring that neither transcribers nor researchers have access to any user-identifiable information.
\end{itemize}

\begin{table}[]
\centering
\caption{Summary of data splits from multiple corpora}
\begin{tabular}{l|ccc}
\hline
\multirow{2}{*}{Domain} & \multicolumn{3}{c}{Splits}                                                                                                                                                                            \\ \cline{2-4} 
                        & \begin{tabular}[c]{@{}c@{}}Text-only\\ (\# text records)\end{tabular} & \begin{tabular}[c]{@{}c@{}}Text-audio\\ (\# utts)\end{tabular} & \begin{tabular}[c]{@{}c@{}}Test\\ (\# utterances)\end{tabular}              \\ \hline
\textsl{conv}                    & 2.9M                                                               & 10K                                                       & \begin{tabular}[c]{@{}c@{}}\textit{conv1}: 3.6K\\ \textit{conv2}: 13K\end{tabular}     \\ \hline
\textsl{wsj}                     & -                                                                  & 10K                                                       & \begin{tabular}[c]{@{}c@{}}\textit{nov93dev}: 0.5K\\ \textit{nov92}: 0.3K\end{tabular} \\ \hline
\textsl{wiki}                    & 1.5M                                                               & -                                                         & -                                                                    \\ \hline
\textsl{video}                   & -                                                                  & 10K                                                       & -                                                                    \\ \hline
\textsl{lib-dev}             & -                                                                  & 5.5K                                                      & -                                                                    \\ \hline
\end{tabular}
\label{table:data_summary}
\end{table}

\subsection{Setups}

The ASR used in our experiments is an RNN-T based model which includes around 80M parameters. It contains an Emformer \cite{emformer2021streaming} based encoder, a predictor with two LSTM layers, and a joiner. The model is trained from scratch on the training split of Librispeech.
All NNLMs in this study are of the same architecture with around 15M parameters, which contains an embedding layer of dimension 300 and two LSTM layers with hidden dimension 1,500. 

We consider the following two domain adaptation tasks in our experiments and introduce corresponding LLMs used in each task.
\begin{itemize}
\item \textbf{Domain adaptation with sufficient text records.} In this case, we have the access of moderate to large amount of text-only records from the target domain. We thus only need to predict the fallibility scores for correction focused NNLM training. We fine-tune RoBERTa with 125M parameters \cite{LiuOyyGoy2019} to serve as fallibility score predictor. 
We perform fine-tuning on all the parameters. 

\item \textbf{Domain adaptation with insufficient text records.} We possess small or zero amount of text-only data from the target domain. In this case, we need to fine-tune LLMs with multi-task training to generate target-domain specific text data along with predicted fallibility scores. We explore an encoder-decoder model T5 \cite{RafShaRob2020}, and a decoder-only model LLaMA2-chat \cite{TouMarSto2023}, which contains 770M and 7B parameters respectively. We conduct full fine-tuning for T5 with fixed prompt as encoder input. Given the large size of LLaMA2-chat, we conduct low-rank adaptation (LoRA) fine-tuning \cite{HuSheWal2022} on all transformer layers.  We also provide few-shot prompts for LLaMA2-chat at both fine-tuning and generation time.
\end{itemize}
For each domain adaptation task, we first acquire the annotated corpus by annotating ASR errors over the text-audio split from the target domain, then fine-tune LLMs for fallibility score prediction and text generation on annotated corpus. Then we conduct correction focused NNLM training on (generated) text-only corpus and integrate NNLMs with ASR through first-pass shallow fusion. The performance is evaluated on test sets from the target domain in terms of perplexity (PPL) and word error rate (WER).

\begin{table}[]
\centering
\caption{Domain adaptation performance on \textsl{conv} test sets}
\begin{tabular}{ll|lr|lr}
\hline
\multicolumn{2}{l|}{\multirow{2}{*}{\begin{tabular}[c]{@{}l@{}}NNLM training\\  strategy\end{tabular}}}             & \multicolumn{2}{c|}{\textit{conv1}} & \multicolumn{2}{c}{\textit{conv2}} \\ \cline{3-6} 
\multicolumn{2}{l|}{}                                                                                               & WER          & PPL         & WER          & PPL        \\ \hline
\multicolumn{2}{l|}{No NNLM}                                                                                        & 26.40         & -$\quad$        & 25.50         & -$\quad$       \\ \hline
\multicolumn{2}{l|}{reg. NNLM ($\alpha=1$)}                                                                                      & 22.92        & 47.3        & 18.40        & 26.1       \\ \hline
\multicolumn{1}{c|}{\multirow{4}{*}{\begin{tabular}[c]{@{}c@{}}Correction \\ focused\\ NNLM\end{tabular}}} & $\alpha$ = 2 & 22.00        & 39.6        & 17.69        & 22.0       \\
\multicolumn{1}{c|}{}                                                                                      & $\alpha$ = 3 & \textbf{21.92}           & \textbf{38.0}        & \textbf{17.46}        & \textbf{20.6}       \\
\multicolumn{1}{c|}{}                                                                                      & $\alpha$ = 5 & 22.76        & 47.9        & 18.30         & 26.7       \\
\multicolumn{1}{c|}{}                                                                                      & $\alpha$ = 8 & 23.13        & 52.7        & 18.60         & 30.8       \\ \hline
\end{tabular}
\label{table: conv_results}
\end{table}

\begin{table}[]
\centering
\caption{Domain adaptation performance on \textsl{wsj} test sets}
\begin{tabular}{ll|lr|rr}
\hline
\multicolumn{2}{l|}{NNLM training strategy} & \multicolumn{2}{c|}{\textit{nov93dev}} & \multicolumn{2}{c}{\textit{nov92}} \\ \hline
\multicolumn{1}{l|}{training corpus}   & $\alpha$  & WER            & PPL          & WER          & PPL        \\ \hline
\multicolumn{1}{l|}{n/a (No NNLM)}     & -  & 19.40          & -$\quad$            & 11.13        & -$\quad$          \\ \hline
\multicolumn{1}{l|}{\textsl{wiki}}              & 1  & 18.18          & 101.0         & 9.35        & 98.6       \\
\multicolumn{1}{l|}{\textsl{T5 gen.}}           & 1  & 17.29          & 89.0         & 8.87        & 78.8       \\
\multicolumn{1}{l|}{\textsl{T5 gen.}}           & 3  & 17.02          & 85.3         & 8.70        & 73.0       \\
\multicolumn{1}{l|}{\textsl{LLaMA2-chat gen.}}  & 1  & 17.10          & 75.0         & 8.12        & 60.3       \\
\multicolumn{1}{l|}{\textsl{LLaMA2-chat gen.}}  & 2  & \textbf{16.00}          & \textbf{73.2}         & \textbf{7.96}        & \textbf{56.7}       \\ \hline
\end{tabular}
\label{table: wsj_results}
\end{table}

\subsection{Results}
Table \ref{table: conv_results} summarizes the results of adaptation task for \textsl{conv} domain, which is considered as text sufficient domain. We compare the results between the regular (reg.) NNLM where all words are treated equally at training time, and the NNLMs trained with correction focused strategy. We explore the impact of $\alpha$ on each test set. 
We observe the best performance when adopting correction focused training with $\alpha = 3$. Comparing with regular NNLM, it achieves relatively 4.3\% to 5.5\% WER reduction on \textsl{conv1} and \textsl{conv2} sets respectively. $\alpha = 2$ achieves similar performance, while $\alpha \ge 5$ may lead to over biasing on words with high fallibility scores and thus undermining the performance.

\begin{figure}
    \centering
    \includegraphics[scale=0.29]{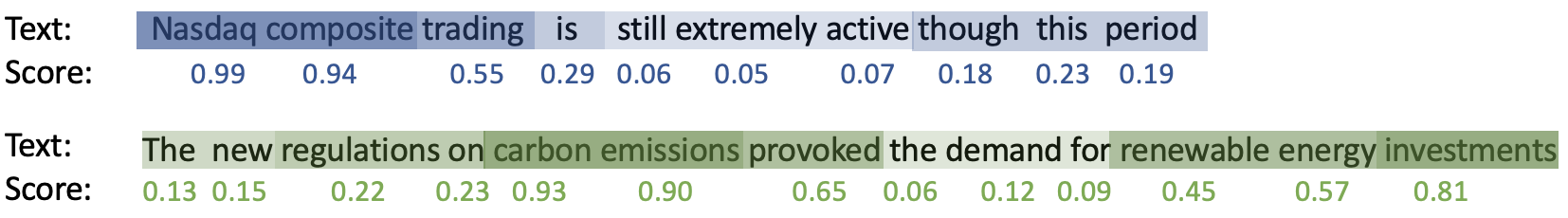}
    \caption{Examples of generated text and associated fallibility scores from T5 (up) and LLaMA2-chat (down) }
    \label{fig:gen_example}
\end{figure}

Table \ref{table: wsj_results} summarizes the results of adaptation task for \textsl{wsj} domain, which is considered as text insufficient domain since we do not have text-only corpus from the target domain. We adopt T5 and LLaMA2-chat to generate 1M target-domain text records with predicted fallibility scores respectively. Examples of generated text with associated fallibility scores are shown in Figure \ref{fig:gen_example}. We use NNLM trained on \textsl{wiki} corpus which contains 1.5M records as the baseline NNLM. We observe relatively 5\% and 13\% WER reduction on \textit{nov92} with NNLMs trained on T5 and LLaMA2-chat generated data respectively. This indicates the high quality and domain fitness of the generated text. Compared with regular NNLM ($\alpha=1$), correction focused NNLMs trained on LLaMA2-chat generated data further reduce WER by up to relatively 6\% WER on \textit{nov93dev} set, however, the PPL on \textit{nov93dev} is not significantly reduced. This further indicates that the correction focused training contributes more on maximizing likelihood for ASR fallible words rather than minimizing the overall PPL.

\subsection{Analysis}

\begin{table}[]
\centering
\caption{Analysis results on \textsl{conv} domain.}
\begin{tabular}{ccc|l|l}
\hline
\multicolumn{3}{c|}{NNLM training strategy}                                       & \multirow{2}{*}{\begin{tabular}[c]{@{}l@{}}\emph{conv1} \\ WER\end{tabular}} & \multirow{2}{*}{\begin{tabular}[c]{@{}l@{}}\emph{conv2} \\ WER\end{tabular}} \\ \cline{1-5}
\multicolumn{1}{c|}{training corpus} & \multicolumn{1}{c|}{annotated corpus} & $\alpha$ &                                                                       &                                                                       \\ \hline
\multicolumn{1}{c|}{\textsl{conv}}            & \multicolumn{1}{c|}{-}                 & 1 & 22.92                                                                 & 18.40                                                                 \\
\multicolumn{1}{c|}{\textsl{conv}}            & \multicolumn{1}{c|}{\textsl{lib-dev}}           & 3 & 22.32                                                                 & \textbf{17.80}                                                                 \\
\multicolumn{1}{c|}{\textsl{conv}}            & \multicolumn{1}{c|}{\textsl{video}}             & 2 & \textbf{22.24}                                                                 & 17.85                                                                 \\ \hline
\multicolumn{1}{c|}{\textsl{wiki}}            & \multicolumn{1}{c|}{-}                 & 1 & 23.42                                                                 & 19.62                                                                 \\
\multicolumn{1}{c|}{\textsl{wiki}}            & \multicolumn{1}{c|}{\textsl{conv}}              & 2 & \textbf{22.68}                                                                 & \textbf{18.50}                                                                  \\ \hline
\end{tabular}
\label{table: analysis}
\end{table}

To verify the effectiveness of fallibility scores, we experiment on \textsl{conv} domain with several variations and summarize the results in Table \ref{table: analysis}. We only fine-tune LLMs as fallibility score predictor and would like to know the impact of domain mismatch between annotated corpus used in LLM fine-tuning and NNLM training corpus. Table \ref{table: conv_results} has shown the effectiveness of predicted fallibility scores. In the case of no annotated corpus available from the target domain, we adopt annotated corpus from other domains which could be the same as (\textsl{lib-dev}) or apart from (\textsl{video}) ASR training domain. We compare the results with the regular NNLM trained on \textsl{conv} corpus without correction focus. We observe similar improvement when fallibility predictor is fine-tuned on \textsl{lib-dev} corpus and \textsl{video} corpus. Considering \textsl{lib-dev} as a smaller annotated corpus than \textsl{video}, the results indicate that the score predictor fine-tuned on ASR training domain corpus can help pick ASR fallible words more precisely. Although the results are not as good as the one adopting target-domain annotated corpus as shown in Table \ref{table: conv_results}, the findings provide options to utilize correction focus training strategy with the absence of target domain annotated corpus. We further investigate the impact of predicting fallibility scores for non-target domain corpus \textsl{wiki} with predictor fine-tuned on target domain annotated corpus. We observe significant improvement in such case indicating the effectiveness of fallibility scores. This provides another option for utilizing non-target domain text corpora in correction focused NNLM training.

\section{Conclusion}
In this work, we propose correction focused LM training strategy to better boost performance on ASR fallible words. Correction focused training relies on word-level ASR fallibility scores which we define as the likelihood of being mis-recognized by a given ASR. We leverage LLMs to assist the data preparation of correction focused training. The LLMs are fine-tuned to serve as fallibility score predictor and text generator simultaneously through multi-task fine-tuning, which significantly improves data preparation efficiency. Experiment results show profound effectiveness of correction focused training indicating the informativeness of fallibility scores.

In the future, we will continue explore this direction by leveraging more information from ASR decoding results such as confusion word pairs.

\vfill\pagebreak


\bibliographystyle{IEEEbib}
\footnotesize
\bibliography{refs}

\end{document}